\documentclass{article}
\usepackage{spconf,amsmath,graphicx,multirow}

\usepackage[
backend=biber,
style=numeric,
sorting=none
]{biblatex}

\newcommand{\wer}{{\footnotesize WER}}
\newcommand{\werr}{{\footnotesize WERR}}

\title{Long-span language modeling for Speech Recognition}

\name{Sarangarajan Parthasarathy\quad William Gale\quad Xie Chen\quad George Polovets\quad Shuangyu Chang}

\address{Microsoft, USA}

\begin{document}
\ninept

\maketitle

\begin{abstract}
We explore neural language modeling for speech recognition where the context spans multiple sentences. Rather than encode history beyond the current sentence using a cache of words or document-level features, we focus our study on the ability of LSTM and Transformer language models to implicitly learn to carry over context across sentence boundaries. We introduce a new architecture that incorporates an attention mechanism into LSTM to combine the benefits of recurrent and attention architectures. We conduct language modeling and speech recognition experiments on the publicly available LibriSpeech corpus. We show that conventional training on a paragraph-level corpus results in significant reductions in perplexity compared to training on a sentence-level corpus. We also describe speech recognition experiments using long-span language models in second-pass re-ranking, and provide insights into the ability of such models to take advantage of context beyond the current sentence.
\end{abstract}


\section{Introduction}
\label{sec:intro}
Language models used in automatic speech recognition (ASR) systems are typically trained on a sentence-level corpus. Intuition suggests that context beyond the current sentence should influence next word prediction. Some efforts at improving the quality of language models using long-span contextual information include incorporating a short-term cache \cite{Kuhn:1990:CNL:80960.80973}, semantic information at the document level \cite{Bellegarda:2000}, dialog states \cite{Xu:2000}, and the use of word-triggers \cite{Lau:1993:TLM:1946943.1946957}. Such long-span context dependencies are difficult to model using n-gram language models (NGLM) commonly used in first-pass ASR decoding. Neural Network Language Models (NNLM) exploit longer context better than NGLMs and have demonstrated significant improvement in perplexity \cite{Mikolov:2000}, and word-error-rate (WER) when used in second-pass rescoring \cite{Liu:2014,Irie2019} and more recently in first-pass decoding \cite{huang2014cache,beck2019lstm}.

While there have been attempts to train NNLMs at the document-level \cite{lin-etal-2015-hierarchical}, NNLMs used in ASR are still trained on a sentence-level corpus. There are many reasons for this. Longer context may not be available or be relevant for improving next-word prediction in commercial ASR systems. For instance, in voice search, long distance word history may be less relevant than non-lexical features such as the geographic location of the user \cite{Biadsy2017EffectivelyBT}. It is also often difficult to obtain training data representing long session contexts in many conversational scenarios. Scenarios where long-span models are useful are becoming more pervasive. Transcriptions of talks and meetings, human-to-human conversation, and document creation by voice, are some scenarios which will likely benefit greatly from long-span models \cite{xiong-etal-2018-session}.

In this work, we restrict our attention to context which consists of word history alone, rather than contexts such as topic of conversation and other non-lexical information. We study the benefits of training NNLMs at the {\it paragraph-level}, where a paragraph is a sequence of consecutive sentences. Rather than summarize the recent past using a word-cache, topic-vectors, etc., we simply concatenate sentences with a sentence boundary symbol which is treated as a word in the vocabulary. We study two popular NNLM architectures, LSTM \cite{Sundermeyer2012LSTMNN}, and Transformer \cite{vaswani2017nips,irie2019interspeech}, and introduce a new variant which augments LSTM with an attention layer. We show that all three architectures are able to take advantage of the longer context to reduce perplexity as well as WER.

\section{Model architectures}
\label{sec:architectures}
We study the following popular architectures for neural language models.

\subsection{Long short-term memory language models}
\label{ssec:lstmlm}

We use a standard LSTM-LM architecture \cite{Jozefowicz:2016}. The modeling unit is a word. The parameters of the input embedding and output linear layers are tied. In order to decouple the choice of the dimensionality of the LSTM layer and the embedding dimension, the output of the final LSTM layer is projected to the embedding dimension. We use noise-contrastive-estimation (NCE) loss during training \cite{Gutmann:2012:NEU:2188385.2188396,dyer2014notes}, which results in approximately self-normalized models. The vocabulary size of our model is 200K which makes training using cross-entropy loss infeasible. Inference, especially in a re-ranking setting, can also be efficiently implemented with self-normalized models, since the linear transform followed by softmax can be replaced by a dot-product. Even though we use NCE loss during training, all perplexity results reported in this paper are computed using cross-entropy to ensure a valid probability distribution.

\subsection{LSTM with attention}
While LSTMs are quite adept at representing history through their hidden states, we hypothesize that in longer sequences, the contributions of earlier words get attenuated. Attention mechanisms on the other hand, are able to assign high weight to any words in the history if they are relevant to the current context. In this work, we add a multi-headed dot product attention layer \cite{vaswani2017nips} over the LSTM hidden states to better utilize the information in long-spanning sequences.

	\begin{equation}
	a_t = \operatorname{MultAttn}(h_1, ..., h_t)
	\end{equation}

	\begin{equation}
	p_t = \operatorname{Proj}(\operatorname{Cat}(a_t, h_t))
	\end{equation}
	
	$\operatorname{MultiAttn}$ is the multi-headed attention function presented in \cite{vaswani2017nips}. It operates over all hidden state outputs at time $t^\prime \leq t$ to produce an attention vector $a_t$. The attention vector is combined with the hidden state output $h_t$ using the concatenation operator $\operatorname{Cat}$ and is passed to the network's output linear layer $\operatorname{Proj}$. 
\label{ssec:lstmattn}

\subsection{Transformer Language model}
\label{ssec:transformer}
In recent years, there is an increasing research endeavor to replace LSTM with transformer \cite{vaswani2017nips} for sequence modeling. Transformer has achieved state-of-the-art in a range of NLP tasks including language models~\cite{devlin2018bert, Irie2019}. In transformer, recurrent connection is not needed.  Self-attention with multi-head is applied to model the long-term history information. 
In this work, we investigate the use of standard transformer architecture for language modeling \cite{vaswani2017nips}, which is similar to the structure used in \cite{radford2018improving}. The relative position embedding was used to model the position information \cite{dai-etal-2019-transformer} for better performance. It is worth noting that, when the number of transformer layers is large (e.g. 10), it is crucial to use warm-up step to increase the learning rate gradually with the progress of training \cite{popel2018training} in order to guarantee the convergence of transformer LMs.

\section{Dataset}
\label{sec:dataset}
We use the publicly available LibriSpeech data-set \cite{Panayotov:2015} in our experiments. Training data for language model consists of a sentence-level corpus of about $803$M words. A vocabulary of $200$K words is also specified. We set up baseline language models on this corpus as distributed, so that we can compare our results to published benchmarks. However, this corpus is not suitable for our experiments with paragraph-level, long-span, language models.

\begin{figure}[htb]
\begin{minipage}[b]{1.0\linewidth}
\centering
\centerline{\includegraphics[width=8.5cm]{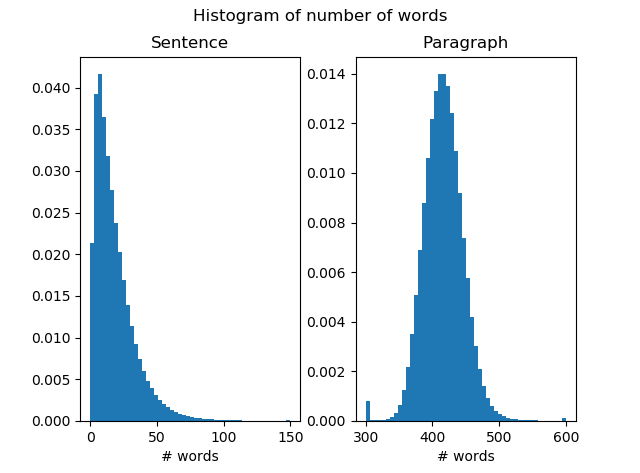}}
\end{minipage}
\caption{Distribution of sentence and paragraph lengths in words in the LM training data.}
\label{figure:1}
\end{figure}

We created a new paragraph-level corpus as follows. We started with the raw text from the same books that were used to created the standard LibriSpeech LM training corpus. We applied the text processing scripts in Kaldi \cite{kaldi_librispeech} to normalize the text exactly as was done to create the standard corpus. We then split the text into sequences of approximately $2000$ characters, taking care to split at the nearest sentence boundary. We call each such sequence, typically containing multiple consecutive sentences separated by a boundary symbol $<$s$>$, a paragraph. As a sanity check of our text processing, we ensured that we could recreate the sentence-level corpus by splitting on sentence boundaries and retaining only unique sentences. Further, by splitting the paragraph-level corpus at sentence boundaries, we get a new sentence-level corpus which is exactly matched in terms of number of words with the paragraph-level corpus. This makes our comparisons of paragraph and sentence models meaningful. This new corpus has about $880$M words. The length distribution of sentences and paragraphs in {\it words} is shown in Fig.~\ref{figure:1}. We similarly created paragraph-level dev and test sets by joining consecutive sentences for evaluating long-span effects.

For word-error-rate (WER) evaluations, we need an acoustic model. Since the focus of this study is long-span language modeling, we did not train an acoustic model on the LibriSpeech audio corpus. For convenience, we took an off-the-shelf acoustic model trained on $1000$s of hours of audio from a variety of Microsoft ASR applications. Therefore, the WER reported in this paper are not directly comparable to WER reported in other publications on LibriSpeech. Since our WERs are in the same ballpark as previously published results, we believe that any conclusions regarding word-error-rate-reductions (WERR) relative to our baseline are still meaningful and should carry-over to other ASR systems.

Finally, we follow common practice \cite{Irie2019} and combine {\it clean} and {\it other} partitions of the dev and test sets for language model evaluations while keeping them separate for WER evaluations.

\section{Experiments}
\label{sec:exp}
\subsection{Language modeling}
\label{ssec:lm}
The goal of these experiments is to study the behavior of LMs trained on a paragraph-level corpus instead of a sentence-level corpus. Therefore, rather than sweep hyper-parameters of the models to get the best possible performance in each scenario, we selected a model size that provides nearly the best performance on this corpus, and kept it constant in all the experiments to make fair comparisons.
\subsubsection{Baseline on standard corpus}
\label{sssec:lmstd}
We trained a $4$-gram NGLM, and a $4\!\times\!2048\!:\!512$ LSTM-LM, on the standard LibriSpeech corpus, where $4$ is the number of layers, $2048$ is the dimensionality of the LSTM state, $512$ is the dimensionality of the embedding and also the output dimensionality of the projection layer. The transformer LM consists of 16 transformer layers, where each transformer layer contains 768 hidden nodes with 12 heads.
Perplexity of the LMs on dev and test sets are shown in Table \ref{table:1}. These perplexities are consistent with best reported results for this models size \cite{Irie2019}.

\begin{table}[h!]
\centering
\begin{tabular}{|l|r|r|}
\hline
Model & dev & test \\
\hline
KN4 & 144.2 & 148.9 \\
LSTM & 62.8 & 65.6 \\
\hline
\end{tabular}
\caption{Perplexity of Kneser-Ney smoothed $4$-gram and LSTM LM on the standard LibriSpeech LM training corpus.}
\label{table:1}
\end{table}

\vspace{-.7cm}
\subsubsection{Baseline on the paragraph corpus}
\label{sssec:lmpara}
We retrained the models in Section \ref{sssec:lmstd} on the sentence and paragraph level corpus created as described in Section \ref{sec:dataset}. Both sentence-level and paragraph-level trained models are evaluated on sentence-level and paragraph-level dev and test sets. Perplexity on {\it sent} evaluation sets of sentence-level NGLM and LSTM-LM are directly comparable to the results in Table \ref{table:1}. The slight difference in the training set causes the perplexities on the same evaluation sets to be higher by about $3$ absolute points for NGLM and about $1$ absolute point for LSTM-LM. This establishes the new baseline for the rest of our experiments.

\begin{table}[h!]
\centering
\begin{tabular}{|l|r|r|r|r|}
\hline
  \multirow{1}{*}{Model} & \multicolumn{2}{c|}{sent} & \multicolumn{2}{c|}{para} \\
   \cline{2-5} \multirow{1}{*}{} & dev & test & dev & test \\
   \hline
   KN4       & 147.4 & 153.8 & & \\
   \hline
   {\small LSTM}-sent & 63.5 & 66.6 & 60.6 & 63.0 \\
   {\small LSTM}-para & 64.4 & 67.6 & 50.3 & 52.1 \\
   \hline
   LSTMA-sent & 62.3 & 65.4 & 79.4 & 83.8 \\
   LSTMA-sent (RA) & & & 64.4 & 67.0 \\
   LSTMA-para & 62.7 & 65.5 & 47.2 & 48.8 \\
   \hline
   Trans-sent & 58.9 & 61.6 & 71.3 & 73.7 \\
   Trans-para & 61.6& 64.1&  48.6 & 50.6 \\
\hline
\end{tabular}
\caption{Perplexity of models trained on sentence and paragraph corpus and evaluated on sentence and paragraph evaluation sets. Trans refers to the transformer model described in Section \ref{ssec:transformer} and LSTMA refers to the LSTM with attention described in Section \ref{ssec:lstmattn}. LSTMA-sent (RA) row uses LSTMA-sent but imposes restriction on the attention-span during inference. }
\label{table:2}
\end{table}

Here are some observations from Table \ref{table:2}. First, consider the LSTM results. Sentence-level models do carry over context across sentence boundaries as evidenced by the lower perplexity in the para columns relative to sent columns. This is in spite of the fact that sentence models have never seen sentence boundaries in the middle of a text sequence during training. Perplexity gains are about $4\%$ relative. The behavior of LSTMA as well as Trans models is different from LSTM models. There is a significant increase in perplexity when sentence models are evaluated on paragraph data. We hypothesized that this is due to mismatch in the time-span over which the attention vector is computed, between training and inference. The mean and standard deviation of the sentence length used to train LSTMA-sent is approximately $19$ and $16$ respectively. We recomputed the perplexity on the paragraph corpus using the sentence-level LSTMA model by restricting the attention-span to 35 past words. The results are shown in row LSTM-sent (RA). Notice that there is no significant drop in perplexity when restricting attention-span in this way. The behavior of paragraph-level models is consistent across all three model architectures. Paragraph-level models perform slightly worse on sentence evaluation data relative to sentence-level models, probably due to mismatch between the lengths of the evaluation and training sequences. The perplexity gains of paragraph-level models on paragraph evaluation data are substantial. If we compare matched perplexities of conventional sentence model evaluated on sentence data and paragraph model on paragraph data, the relative gains are about $20\%$, $24\%$, and $17\%$, for LSTM, LSTMA, and Trans architectures respectively.

\subsection{Speech recognition: n-best re-ranking}
\label{ssec:asr}
As mentioned in Section \ref{sec:dataset}, we use an off-the-shelf acoustic model and the $4$-gram sentence-level NGLM, to generate n-best hypotheses using a WFST ASR decoder. We then re-rank the n-best hypotheses using various NNLMs, using log-linear combination of AM, first-pass NGLM, and NNLM likelihoods. It is important to point out that the first-pass recognizer still operates at the sentence-level and there is no state carry-over across sentence boundaries. Therefore, we are inherently limited by how much context information can be injected into second-pass re-ranking. The word-error-rate of the top-choice hypothesis of first-pass ASR is shown in Table~\ref{table:3}.
\begin{table}[h!]
\centering
\begin{tabular}{|l|r|}
\hline
  eval set & first-pass \wer \\
  \hline
  dev-clean & 4.63 \\
  dev-other & 10.67 \\
  test-clean & 4.84 \\
  test-other & 10.94 \\
\hline
\end{tabular}
\caption{1-best \wer in \% of the first-pass decoder}
\label{table:3}
\end{table}

The first set of experiments demonstrate the benefits of re-ranking using NNLMs using traditional sentence-level models. The re-ranked top-choice \wer as well as relative WER reduction (\werr) using three NNLMs described in Section \ref{sec:architectures}, are shown in Table \ref{table:4}. The corresponding results using paragraph LMs are shown in Table \ref{table:5}. It is clear that re-ranking using NNLMs is effective in reducing \wer. All three models perform roughly similarly. The Transformer LM performs the best by a slight margin. The performance of paragraph LMs is also very close to the performance of sentence LMs since there is no additional context presented to the model.

\begin{table}[h!]
\centering
\begin{tabular}{|l|r|r|r|r|r|r|}
\hline
  \multirow{1}{*}{eval set} & \multicolumn{2}{c|}{LSTM-sent} & \multicolumn{2}{c|}{LSTMA-sent} & \multicolumn{2}{c|}{Trans-sent} \\
  \cline{2-7}
  \multirow{1}{*}{} & \wer & \werr & \wer & \werr & \wer & \werr \\
  \hline
  dev-clean & 2.91  & 37.13 & 2.91 & 37.09 & 2.78 & 39.87 \\
  dev-other & 7.46  & 30.09 & 7.44 & 30.29 & 7.24 & 32.13 \\
  test-clean & 3.16 & 34.81 & 3.24 & 33.12 & 3.01 & 37.83 \\
  test-other & 7.68 & 29.76 & 7.58 & 30.72 & 7.44 & 32.01 \\
  \hline
\end{tabular}
\caption{Re-ranked 1-best \wer(\%) and \werr(\%) using sentence-level NNLMs on sentence data}
\label{table:4}
\end{table}

\begin{table}[h!]
\centering
\begin{tabular}{|l|r|r|r|r|r|r|}
\hline
  \multirow{1}{*}{eval set} & \multicolumn{2}{c|}{LSTM-para} & \multicolumn{2}{c|}{LSTMA-para} & \multicolumn{2}{c|}{Trans-para} \\
  \cline{2-7}
  \multirow{1}{*}{} & \wer & \werr & \wer & \werr & \wer & \werr \\
  \hline
  dev-clean & 2.98 & 35.58  & 2.87 & 37.92 & 2.79 & 39.63 \\
  dev-other & 7.53 & 29.39  & 7.49 & 29.80 & 7.35 & 31.08 \\
  test-clean & 3.21 & 33.67 & 3.15 & 34.97 & 3.03 & 37.40 \\
  test-other & 7.67 & 29.90 & 7.53 & 31.10 & 7.43 & 32.08 \\
  \hline
\end{tabular}
\caption{Re-ranked 1-best \wer(\%) and \werr(\%) using paragraph-level NNLMs on sentence data.}
\label{table:5}
\end{table}

In order to determine the extent to which paragraph models can take advantage of additional context, we did a cheating experiment where we scored the current sentence in the context of the past sentences within a paragraph. We used the reference transcripts rather than the recognized hypotheses to understand the ability of the NNLMs to use past context. The results are shown in Table \ref{table:6}. Improvement in WERR over sentence models is consistent across all data sets. The two conditions we care about are the performance of the conventional re-ranker using sentence-models in sentence context (S/S), and the new paragraph models evaluated in paragraph context (P/P). First-pass ASR still decodes only in sentence context. The WERR of P/P relative to S/S averaged across the data-sets is $6.6\%$, $9.2\%$, and $4.2\%$ for LSTM, LSTMA, and Trans models respectively. While all architectures are able to take advantage of the paragraph context, LSTMA architecture seems to achieve the largest gains. The somewhat lower gains of the Transformer model due to just longer LM context may be because the absolute WER of S/S system is still lower than the other two architectures and search error in the n-best hypotheses may be limiting the gains. We intend to implement a second-pass lattice decoder to fully take advantage of the perplexity gains offered by longer context shown in Table~\ref{table:2}.

\begin{table}[h!]
\centering
\begin{tabular}{|l|r|r|r|r|r|r|}
\hline
  \multirow{1}{*}{eval set} & \multicolumn{2}{c|}{LSTM-para} & \multicolumn{2}{c|}{LSTMA-para} & \multicolumn{2}{c|}{Trans-para} \\
  \cline{2-7}
  \multirow{1}{*}{} & \wer & \werr & \wer & \werr & \wer & \werr \\
  \hline
  dev-clean & 2.71  & 41.45 & 2.59 & 43.99 & 2.64 & 43.00 \\
  dev-other & 7.06  & 33.82 & 6.87 & 35.60 & 6.92 & 35.17 \\
  test-clean & 2.97 & 38.65 & 2.90 & 40.18 & 2.99 & 38.34 \\
  test-other & 7.06 & 35.41 & 6.99 & 36.11 & 6.98 & 36.18 \\
  \hline
\end{tabular}
\caption{Re-ranked 1-best \wer(\%) and \werr(\%) using paragraph-level NNLMs scored in {\it reference} paragraph context.}
\label{table:6}
\end{table}

In practice, first-pass ASR generates a sequence of n-best sentence-level hypotheses in each session. To obtain paragraph-level LM scores for the current sentence, we need to determine which of the contexts from the previous n-best hypotheses to carry over. This would typically be implemented using a beam-search. We tried a simple strategy where we carry-over only the context of the 1-best hypothesis from the previous sentence. The results are shown in Table \ref{table:7}. This simple strategy for context carry-over is effective in achieving most of the potential gains shown in the cheating results in Table \ref{table:6}.
\begin{table}[h!]
\centering
\begin{tabular}{|l|r|r|}
\hline
  \multirow{1}{*}{eval set} & \multicolumn{2}{c|}{LSTM-para} \\
  \cline{2-3}
  \multirow{1}{*}{} & \wer & \werr  \\
  \hline
  dev-clean & 2.80  & 39.43  \\
  dev-other & 7.24  & 32.17  \\
  test-clean & 2.97 & 38.65  \\
  test-other & 7.27 & 33.54  \\
  \hline
\end{tabular}
\caption{Re-ranked 1-best \wer(\%) and \werr(\%) using paragraph-level NNLMs scored in {\it 1-best hypothesis} paragraph context.}
\label{table:7}
\end{table}

Since our interest is primarily in studying the effect of longer context in language modeling, we have fixed the ASR system and the re-ranker parameters to reasonable settings but not tuned them. While such tuning may lower \wer, it is unlikely to change our conclusions about the benefits of using longer context. In order to disentangle the ASR effects further, we simulated a situation where we eliminated search error by adding the reference transcription to the n-best list used for re-ranking. This involved generating AM likelihoods using forced-alignment of the reference transcription with the audio, and computing the first-pass NGLM likelihood for the reference. Just to be certain that the behavior of the ASR decoder in forced-alignment mode is not subtly different than during normal decoding, we also generated the AM and first-pass LM likelihoods for the n-best hypotheses using forced-alignment of each hypothesis with the audio. The results are shown in Table \ref{table:8}. The baseline WER in Table \ref{table:8} is different from the one in Table \ref{table:3}. The main reason is out-of-vocabulary words in the reference transcription (OOV). For example, if sentences with OOV are removed from scoring, the WER for dev-clean drops from $4.63$ to $4.04$. The rest of the drop to $3.86$ is explained by search error. The OOV filtering study was only for diagnosis purposes. The results reported in Table \ref{table:8} use exactly the dev and test sets used in other experiments.

\begin{table}[h!]
\centering
\begin{tabular}{|l|r|r|r|r|r|}
\hline
  \multirow{1}{*}{eval condition} & \multirow{1}{*}{baseline} & \multicolumn{2}{c|}{LSTM-sent} & \multicolumn{2}{c|}{LSTM-para} \\
  \cline{3-6}
  \multirow{1}{*}{} & \wer & \wer & \werr & \wer & \werr  \\
  \hline
  dev-clean (R) & \multirow{2}{*}{3.86} & 2.24 & 41.87 & 2.31 & 40.11  \\
  dev-clean (R+C) &  \multirow{1}{*}{} & 2.23 & 42.16 & 2.01 & 47.96 \\
  \hline
  dev-other (R) & \multirow{2}{*}{9.47} & 5.72 & 39.64 & 5.68 & 40.06  \\
  dev-other (R+C) &  \multirow{1}{*}{} & 5.51 & 41.88 & 4.79 & 49.46 \\
  \hline
\end{tabular}
\caption{Re-ranked 1-best \wer(\%) and \werr(\%) using sentence-level and paragraph-level NNLMs assuming no search error. (R) refers to the condition where the reference transcription is added to the n-best hypotheses from the first-pass. (R+C) refers to the condition where the reference transcription is added to the n-best {\it and} the NNLM scoring uses context from previous sentences in the session. }
\label{table:8}
\end{table}

The most interesting observation from Table \ref{table:8} is that LSTM-para is much more effective at taking advantage of context beyond the current sentence. For example, WER with paragraph-context for LSTM-para is $4.79$ relative to $5.68$ without beyond-sentence context. There is some drop in WER even for LSTM-sent for dev-other but not as significant as for LSTM-para.

\section{Related work}
\label{sec:related}
There has been significant work on long-span neural LMs in the larger NLP community beyond speech recognition \cite{dai-etal-2019-transformer,khandelwal-etal-2018-sharp}. Much of the recent NNLM research treats the entire corpus as a single long string of text and segment it into sequences without regard to sentence boundary. When presented with a evaluation corpus consisting of a collection of sentences or paragraphs, the order of evaluation affects the likelihood of each text string and hence the corpus perplexity calculation. In speech recognition, we have typically insisted on deterministic behavior where the LM likelihood of a sentence is not affected by previous context since the context is reset at the beginning of a sentence. In this work, we still reset the state of our NNLMs, except at the paragraph-level instead of a sentence-level. As we have demonstrated, such a model is effective at traditional sentence-level scoring when no context information is available, and yet take advantage of the longer context when it is available.

Long-span language modeling ideas explored in the context of speech recognition are more relevant to our work. There have been prior attempts at incorporating context beyond the current sentence. A conversational LM that conditioned next-word predictions of a current sentence on words uttered by a different speaker in the conversation was introduced in \cite{Ji:2004:MLM:1613984.1614018}. Cache-based models, where a model trained on the recent past is interpolated with the base LM, have a long history \cite{Kuhn:1990:CNL:80960.80973}. Exponential and trigger-based language models also allow mechanisms for injecting long-distance information \cite{Lau:1993:TLM:1946943.1946957}. A mechanism for injecting a context-vector such as topics in an NNLM was introduced in \cite{Mikolov2012ContextDR}. All of these efforts attempt to explicitly inject long-span information that may be relevant to next-word prediction. These approaches can complement and enhance our models and will be the subject of future work. The work that is most closely related to ours is the work on session-level language modeling for conversational speech \cite{xiong-etal-2018-session}. They train a traditional sentence-level LSTM-LM and score the current hypothesis in the context of words in the previous utterance. Our study is focused on how well NNLMs learn long-span context implicitly when simply presented with past text during training and evaluation. This effort was motivated by promising results we achieved in an earlier unpublished work \cite{Jurik:2018} on an independent text corpus.

\section{Conclusion}
\label{sec:conclusion}
Language models for speech recognition are traditionally trained on sentence-level data. We have demonstrated that LSTM, LSTMA, and Transformer language models, trained and evaluated on paragraph-level data, achieves perplexity reduction of about $20\%$. Such models can be effectively trained without any additional architectural changes to the models, or significant changes to the training methodology. We also demonstrated gains in WERR of $4\% - 9\%$, that can be attributed only to the use of context beyond the current sentence, when evaluating the paragraph-level models.

\clearpage
\printbibliography

\end{document}